\documentclass[onecolumn]{IEEEtran}
\usepackage{cite}
\usepackage{hyperref}
\usepackage{pgfplots}
\pgfplotsset{compat=1.5}
\usepackage{multirow}

\usepackage{amsmath,amsfonts}
\usepackage{enumerate} 

\usepackage{algorithm} 
\usepackage{algorithmic}  

\usepackage{tabularx}
\usepackage{lipsum}

\begin{document}

\title{Multi-objective Differential Evolution with  Helper Functions for Constrained Optimization}

\author{Tao~Xu and Jun~He  
\thanks{This work was supported by the EPSRC under Grant No. EP/I009809/1.}
\thanks{Tao Xu  and Jun He are with Department of Computer Science, Aberystwyth University, Aberystwyth, SY23 3DB, UK. Email:  tax2@aber.ac.uk,   jqh@aber.ac.uk} 
}

\maketitle

\begin{abstract}
Solving constrained optimization problems by multi-objective evolutionary algorithms has scored tremendous achievements in the last decade.  Standard multi-objective schemes usually aim  at minimizing the objective function and also the degree of constraint violation simultaneously. 
This paper proposes  a new multi-objective method for solving constrained optimization problems. 
The new method  keeps two standard    objectives:  the original objective function and   the sum of degrees of constraint violation. But besides them, four more objectives are added. One is based on the feasible rule. The   other three come from the penalty functions.   This paper conducts an initial experimental study on thirteen benchmark functions. A simplified version of CMODE  is applied to solving   multi-objective optimization problems. Our initial experimental results   confirm  our expectation that adding more helper functions could be useful. The performance of SMODE with more helper functions (four or six) is better than that  with only two helper  functions. 
\end{abstract}

%\begin{IEEEkeywords}
%	constrained optimization problems, differential evolution, helper functions, multi-objective optimization.
%\end{IEEEkeywords}

\IEEEpeerreviewmaketitle

\section{Introduction}
Optimization problems in real-world applications are usually subject to different kinds of constraints. These problem are called constraint optimization problems (COPs). In the minimization case,  the COP is formulated  as follows:
\begin{align}
\label{COP}
&\min  f(\vec{x}),  \quad \vec{x}=(x_1,\cdots,x_n) \in S,\\
&\text{subject to} \begin{cases}   g_i(\vec{x})\leq0,\quad    i=1,\cdots,q,\\
 h_j(\vec{x})=0,\quad j=1,\cdots, r,
 \end{cases}
 \end{align}
where  $S$ is a bounded domain in $\mathbb{R}^n$, given by
\begin{equation}
S=\{ \vec{x}\mid  L_i \leq x_i\leq U_i,  i =1, \cdots, n\},
\end{equation} where $ L_i $ is the  lower boundary and $ U_i $  the upper boundary.
$ g_i(\vec{x})\le 0$ is the $i$th inequality constraint while $ h_j(\vec{x})=0$ is the $ j $th equality constraints.
The feasible region $ \Omega\subseteq S $ is defined as:
\begin{align*}
 \{\vec{x}\in S\mid  g_i(\vec{x}) \leq 0, i=1,\cdots,q; h_j(\vec{x})=0, j=1,\cdots,r \}.  
\end{align*}

If an inequality constraint  meets $ g_i(\vec{x})=0 $ (where $i =1,\cdots,q)$ at any point $ \vec{x} \subseteq \Omega$, we say it is active at $ \vec{x}$. All equality constraints $ h_j  (\vec{x})$ (where  $j= 1,\cdots,r)$ are considered active at all points of $\Omega$.

Many constraint-handling techniques have been proposed in literature. The most popular constraint-handling techniques include  penalty function methods, the feasibility rule, multi-objective optimization and repair methods.  A detailed introduction to this topic can be found in several comprehensive surveys \cite{michalewicz1996evolutionary,coello2002theoretical,mezura2011constraint}.

This paper focuses on multi-objective optimization methods, which  are regarded as one of  the most promising ways for dealing with COPs  \cite{segura2013using}. The technique is based on using multi-objective optimization evolutionary algorithms (MOEAs) for solving single-objective optimization problems. This idea can be traced back to 1990s~\cite{louis1993pareto} and it  is also termed multi-objectivization \cite{knowles2001reducing}. Multi-objective methods separate the objective function and the
constraint violation degrees into different fitness functions. This is unlike penalty functions, which combine  them into a single fitness function.  
The main purpose of using multi-objective optimization is to relax the requirement of setting or fine-tuning parameters,  as happens with   penalty function methods. 

 The research of dealing with   COPs using MOEAs has made significant achievements since 2000. There exist variant methods of    applying  MOEAs for solving COPs. According to the taxonomy proposed in~\cite{mezura2008constrained,segura2013using}, these methods are classified into five categories:

\begin{enumerate}
\item \emph{Bi-objective feasible complaints methods:} methods that transform the original single-objective COP   into an unconstrained bi-objective optimization problem, where the first objective is the original  objective function and the second objective is a measure of the constraint violations. During solving the multi-objective problem, selection always prefers a feasible solution over an infeasible solution. \cite{wang2005preference,wang2007multiobjective} are two examples of bi-objective feasible complaints methods. However, the number of research in this category is very limited. 

\item \emph{Bi-objective non-feasible complaints methods:} like the first category, the original single-objective COP   into an unconstrained bi-objective optimization problem. But during solving the latter problem, selection is designed based on the dominance relation and doesn't prefer a feasible solution over an infeasible solution.
A lot of work belong to this category,  such as~\cite{surry1997comoga,zhou2003multi,cai2006multiobjective,
deb2007hybrid,venkatraman2005generic,ray2009infeasibility, jain2014evolutionary,martinez2014multi,gao2015dual}. 

\item \emph{Multi-objective feasible complaints methods:} methods that transform   the original single-objective COP   into an unconstrained multi-objective optimization problem, which includes $1+q+r$ objectives. The first objective is the original objective. The other $q+r$ objectives correspond to each constraint in the COP.  During solving the multi-objective problem, selection always prefer a feasible solution over an infeasible solution.  The work in this category includes \cite{coello2002handling,jimenez2002evolutionary,kukkonen2006constrained,gong2008multiobjective}.

\item \emph{Multi-objective non-feasible complaints methods:} like the third category, the original single-objective COP is transformed  into an unconstrained multi-objective optimization.  But during solving the multi-objective problem, selection doesn't prefer a feasible solution over an infeasible solution. The idea was used in~\cite{ray2000evolutionary,aguirre2004handling,liang2006dynamic}.

\item \emph{Other  multi-objective methods:} methods   that transform   the original single-objective COP   into an unconstrained multi-objective optimization problem, but some or all of the objectives in the latter problem are different from the original  objective function and the degrees of the constraint violation. For example, the first objective in \cite{watanabe2005multi} is the original objective function with addition of noise.  The second objective  equals to the original objective function but considering relaxed constraints. This category is less studied than others. The main problem is how to construction helpful objectives.
\end{enumerate}

The multi-objective method in this paper belongs to the fifth category. Our method still keeps the standard    objectives:  the objective function and also the total degree of constraint violation. But besides them, more objectives are added. One is based on the feasible rule. The others are from the penalty functions. In this way a new multi-objective   model is constructed for constrained optimization. A natural question is to investigate whether adding more objectives can  improve the performance of MOEAs for solving constrained optimization problems.  This paper conducts an experimental study. A simplified version of CMODE~\cite{wang2012combining} is applied to solving for  multi-objective optimization problems. Our initial experimental result is positive. It confirms our expectation that adding helper functions could be useful.

The rest of paper is organized as follows. Section~\ref{secDE} reviews differential evolution. Section~\ref{secMO}  proposes a new multi-objective model for constrained optimization.  Section~\ref{secSMODE} describes a multi-objective differential evolution algorithm with helper functions. Section~\ref{secExperiment} gives experiment results and compares the proposed approach    with different numbers of helper functions. Section~\ref{secConclusion} concludes the paper.

\section{Differential Evolution}
\label{secDE}
Differential evolution (DE)  was proposed by Storn and Price~\cite{storn1997differential}, which is arguably one of the most powerful stochastic real-parameter optimization algorithms in current use~\cite{das2011differential}.

In DE,  a population $P_t$  is represented by   $\mu$ $n$-dimensional vectors:
\begin{align}
& P_t = \{ \vec{x}_{1,t}, \cdots, \vec{x}_{\mu,t} \},\\
& \vec{x}_{i,t}=( x_{i,1,t}, x_{i,2,t}, \cdots, x_{i,n,t}),  i=1, 2, \cdots, \mu,
\end{align}
where $t$ represents the generation counter. Population size $\mu$ does not change during the minimization
process. The initial vectors are  chosen randomly from $[L_i, U_i]^n$. The formula below shows how to generate an initial individual $\vec{x}=( x_1, \cdots, x_n) $ at random:
\begin{equation}
x_i= L_i+(U_i-L_i) \times rand, \quad  i =1, \cdots, n,
\end{equation}
 where $rand$ is the random number $[0,1]$. 

There exist several variants of DE. The original DE algorithm~\cite{storn1997differential} is utilized in this paper.  
This DE algorithm consists of three operations: mutation, crossover and selection,  which are described as follows. 
\begin{itemize}

\item \emph{Mutation:}
for each target  $ \vec{x}_{i,t}$ where $i=1, \cdots, n,$ a mutant vector 
$  \vec{v}_{i,t}=(v_{i,1,t},v_{i,2,t},\cdots,v_{i,n,t}) 
$  is generated by
\begin{equation}
\vec{v}_{i,t} =\vec{x}_{r1,t}+F\cdot( \vec{x}_{r2,t}- \vec{x}_{r3,t})
\end{equation} 
where random indexes $ r1, r2, r3 \in \{1, \cdots, \mu\}$ are mutually different  integers. They are   also chosen to be different
from the running index $i$.  $F$  is a real and constant factor  from $[0, 2]$ which controls the amplification of the differential variation $( \vec{x}_{r2,t}- \vec{x}_{r3,t})$.
In case $\vec{v}_{i,t}$ is out of the interval $[L_i, U_i]$, the mutation operation is repeated until $\vec{v}_{i,t}$ falls in $[L_i, U_i]$.

\item \emph{Crossover:} in order to increase population diversity, crossover is also used in DE.  The trial vector $\vec{u}_{i,t}$ is generated by  mixing the target vector $\vec{x}_{i,t}$ with the mutant vector $\vec{v}_{i,t}$. Trial vector $
 \vec{u}_{i,t} = (u_{i,1,t},  u_{i,2,t}, \cdots,  u_{i,n,t}) 
$
is constructed as follows:
\begin{align}
u_{i,j,t}=&
\begin{cases}
{v}_{i,j,t},      &  \text{if } rand_j(1,0)\leq Cr    \text{ or } j=j_{rand}, \\
{x}_{i,j,t},  &   \text{ otherwise}, 
\end{cases} \nonumber \\
& j=1, \cdots, n, 
\end{align}
where $rand_j(0,1) $ is a uniform random number   from $ [0, 1] $. Index $ j_{rand} $ is   randomly chosen   from $ \{1, \cdots, n \}$.   $Cr\in[0,1] $ denotes the crossover   constant which has to be determined
by the user. In addition, the condition ``$ j=j_{rand} $'' is used to ensure the trial vector $ \vec{u}_{i,t} $ gets at least one parameter from vector $ \vec{v}_{i,t} $.

\item \emph{Selection:}
a greedy criterion is used to  decide which offspring generated by mutation and crossover should be selected to population $P_{t+1} $.  Trail vector $ \vec{u}_{i,t} $ is compared to target vector $ \vec{x}_{i,t} $, then the better one will be reserved to the next generation.

\end{itemize}

\section{Multi-objective Model with More Helper Functions for Constrained Optimization}
\label{secMO}
Without loss of generality, consider a minimization problem with only two constraints:
\begin{align}
\label{equCOP}
\begin{cases}
\min f(\vec{x}), \\
\text{subject to }
g(\vec{x})\leq0 \text{ and } h(\vec{x})=0.
\end{cases}
\end{align}

A multi-objective method transfers the above single-objective optimization problem with constraints into a multi-objective optimization problem without constraints.  

The first fitness function is the original objective function $f(\vec{x})$ without considering constraints:
\begin{align} 
  f_1(\vec{x})=f(\vec{x}).
 \end{align} 
Notice that the optimal solution to minimizing $f_1(\vec{x})$   might be different from that to the original constrained optimization problem~(\ref{equCOP}), therefore $f_1(\vec{x})$ is only a   helper fitness function. 

The second  objective is related to constraint violation. Define the degree of violating each constraint  as
\begin{align}
&v_{1}(\vec{x})= \max \{ 0, g(\vec{x}) \},\\
&v_2(\vec{x})= \max \{0, \lvert h(\vec{x})\rvert-\delta\},
\end{align}
where $\delta$ is   the tolerance allowed for the equality constraint.

The second fitness function is defined by the sum  of constraint violation degrees:
\begin{align}
f_2(\vec{x})=v(\vec{x})=v_1(\vec{x})+v_2(\vec{x}).
\end{align} 
 
The above two objectives are widely used in in multi-objective methods for constrained optimization~\cite{segura2013using}. An interesting question is whether using more fitness fitness function can improve the performance of MOEAs? This paper aims to   investigate the relationship between the performance of multi-objective  and the number of  objectives used. 

A problem is how to construct new helper functions. This paper designs two types of general purpose fitness functions, which are constructed from the feasible rule and the penalty method. Any problem-specific knowledge can be used in designing helper functions. For example, inspired from a greedy algorithm,  several helper functions are specially constructed for solving the 0-1 knapsack problem in\cite{he2014theoretical}.

Besides the original objective function $f_1(\vec{x})$ and the sum of constraint violation degrees $f_2(\vec{x})$, the third fitness function is designed by the feasible rule~\cite{deb2000efficient}. During pairwise-comparing individuals:
\begin{enumerate}
\item when two feasible solutions are compared, the one with a better objective function profit is chosen; 

\item  when one feasible solution and one infeasible solution are compared, the feasible solution is chosen; 

\item  when two infeasible solutions are compared, the one with smaller constraint violation is chosen.
\end{enumerate}  

According to the feasible rule, the third fitness function  is constructed as follows: for an individual $x$ in a population $P$,   
  \begin{align}
 f_3(\vec{x})=
 \left\{
 \begin{array}{lll}
 f(\vec{x}), &\mbox{if } \vec{x} \mbox{ is  feasible};\\
f^{\sharp}+v (\vec{x}) , &\mbox{otherwise}. 
 \end{array}
 \right.
\end{align}  
In the above, $f^{\sharp}$ is  the ``worst'' fitness of  feasible individuals in population $X$, given by
 \begin{align}
f^\sharp=
 \left\{
 \begin{array}{lll}
 \max\{f(\vec{x}); \vec{x} \in P \mbox{ and } \vec{x} \mbox{ is feasible } \};\\
0, \quad \mbox{otherwise}. 
 \end{array}
 \right.
\end{align}
 
Since   the reference point $f^{\sharp}$ depends on    population $P$, thus  for the same $x$,  the values  of $f_3(\vec{x})$ in different populations  $P$ might be different. However the optimal feasible solution to minimizing $f_3(\vec{x})$ always is  the best in any population. Thus the optimal feasible solution  to minimizing $f_3(\vec{x})$ is exactly the same as that to the constrained optimization problem. Based on this reason, $f_3(\vec{x})$ is called an equivalent fitness function.

Inspired from the penalty function method,  more fitness functions with different penalty coefficients are constructed as follows:
\begin{align}
f_4(\vec{x})=f(\vec{x})+c_4 v(\vec{x}), \\
f_5(\vec{x})=f(\vec{x})+ c_5 v(\vec{x}),\\
f_6(\vec{x})=f(\vec{x})+c_6 v(\vec{x}).  
\end{align}
where $c_4, c_5,c_6$ are  penalty coefficient. If set $c_i=+\infty$, then $f_i(\vec{x})$ represents a death penalty to infeasible solutions.
Such a function $f_i(\vec{x})$ is a helper function because minimizing $f_i(\vec{x})$ might not lead to the optimal feasible solution.

In summary, the original constrained optimization problem is transferred into a multi-objective optimization problem:
\begin{align}
\label{equMOP} 
\min (f_1(\vec{x}), \cdots, f_6(\vec{x})),   
\end{align}
which  consists of one equivalent function $f_3(\vec{x})$ and five helper functions. This  new multi-objective model for constrained optimization is the main contribution of this paper. The model potentially  may include   many objectives inside.

\section{Multi-objective Differential Evolution for Constrained Optimization}
 \label{secSMODE}
The  CMODE framework~\cite{wang2012combining} is chosen to solve the above multi-objective optimization problem~(\ref{equMOP}). Different from normal MOEAs, CMODE is specially designed  for solving  constrained optimization problems. Hence it is expected that CMODE is  efficient in solving the   multi-objective optimization problem~(\ref{equMOP}).  A comparison study of several MOEAs is still undergoing.

CMODE~\cite{wang2012combining} originally is applied to solving a bi-objective optimization problem which consists of only two objectives: $f_1(\vec{x})$ and $f_2(\vec{x})$. However, it is easy to reuse the existing framework of CMODE to multi-objective optimization problems. Due to time limitation, a simplified  CMODE  algorithm is implemented in this paper. In order to distinguish the original CMODE, the simplified version is abbreviated by SMODE. The algorithm is described as follows. %in~\ref{algFramwork}.     
 
%\begin{algorithm}[ht]  
%	\caption{Framework of CMODE}
%	\label{algFramwork}  
	\begin{algorithmic}[1]  
		\REQUIRE   
		$\mu $:  population size;\\  
		$\lambda$: the number of individuals involved in DE operations \\  
		  $FES_{\max}$: the maximum number of fitness evaluations
	    	\STATE randomly generate an initial population $P_0$ with population size $ \mu $;
	    	\STATE evaluate the values of $f$ and $v$  for each individual in the initial population, and then calculate the value of $f_i$ where $i =1,\cdots, m$;  
        	\STATE set $FES=\mu$; // $FES$ denotes the number of fitness evaluations;
 %\STATE set $A=\emptyset$; //$A$ an archive to store the infeasible individual with the lowest degree of constraint violation;
        	\FOR{$t=1, \cdots, FES_{\max}$}
        	 \STATE choose $\lambda$ individuals (denoted by $Q$) from population $P_t$;
        	 \STATE let $P'=P_t \setminus Q$;
         \STATE for each individual in set $Q$, an offspring is generated by using DE mutation and crossover operations   as explained in Section~\ref{secDE}. Then $\lambda$ children (denoted by $C$) are generated from $Q$;
     	\STATE evaluate the values of $f$ and $v$   for each individual in   $C$ and then obtain the value of $f_i$ where $i =1,\cdots, m$;  
         	    	
     	        	\STATE set $FES=FES+\lambda$;
\STATE identify all nondominated individuals  in  $C$ (denoted by $R$);
\FOR{each individual $\vec{x}$ in $R$}
\STATE find all individual(s) in $Q$ dominated by $\vec{x}$; 
\STATE randomly replace one of these dominated individuals by $\vec{x}$;
\ENDFOR    	        	
\STATE let $P_{t+1}=P' \cup Q$;
%\IF{no feasible solution exists in $R$}
%\STATE identify the infeasible solution $\vec{x}$ in $R$ with the lowest degree of constraint violation and add $\vec{x} $ to $A$;
%\ENDIF
%\IF{$\mod (t,k)=0$} 
%\STATE execute the infeasible solution replacement mechanism and set $A=\emptyset$; // this mechanism is given in Algorithm~\ref{algMechanism }
%\ENDIF
        	\ENDFOR   
 	    \RETURN  the best found solution  
\end{algorithmic}    
%\end{algorithm}  

The algorithm is explained step-by-step in the following. At the beginning,  an initial population $P_0 $ is chosen at random, where all initial vectors are  chosen randomly from $[L_i, U_i]^n$. 
 
At each generation,  parent population $P_t$ is split into two groups: one group with $\lambda$ parent individuals that  are used for DE operations (set $Q$) and  the other group (set $P'$) with $\mu-\lambda$ individuals that are not involved in DE operations.  DE operations are applied to $\lambda$ selected children (set $Q$) and then generate $\lambda$ children (set $C$).  

Selection is based on the dominance relation. First nondominated individuals (set $R$) are identified from children population $C$. Then these individual(s) will replace the dominated individuals in $Q$ (if exists). As a result, population set $Q$ is updated. Merge population set $Q$ with those parent individuals that are involved in DE operation (set $P'$) together and form the next parent population $P_{t+1}$.     
The procedure repeats until  reaching the maximum number of evaluations. The output is the best found solution    by DE.  

Due to time limitation,  our algorithm doesn't implement a special mechanism used in CMODE: the infeasible solution replacement mechanism. The idea of this replacement mechanism is that, provided that a children population is composed of only infeasible
individuals, the ``best'' child, who has the lowest degree of constraint violation, is stored into an archive. After a fixed interval of generations, some randomly selected infeasible individuals in the archive will replace the same number of randomly selected individuals in the parent population. Although  this significantly influence s the efficiency of our algorithm, our study is still meaningful since our goal is to investigate whether using more objectives may improve the performance of MOEAs for constrained optimization.

\section{Experiments and Results}
\label{secExperiment}

\subsection{Experimental Settings}
In order to study the relationship between the performance of SMODE and the number of helper functions,  thirteen benchmark functions were employed as the instances to perform  experiments. These benchmarks have been used to test the performance of MOEAs for constrained optimization in \cite{cai2006multiobjective} and are a part of benchmark collections in IEEE CEC 2006 special session on constrained real-parameter
optimization~\cite{liang2006problem}. Their
detailed information is provided in Table \ref{tableBenchmark}, where $n$ is the number
of decision variables, $LI$ stands for the number of linear inequalities constraints, $NE$
  the number of nonlinear equality constraints, $NI$ nonlinear inequalities constraints. $\rho $ denotes the ratio between the sizes of the entire search space and feasible search space and $a$ is the number of active constraints at the optimal solution.

\begin{table}[ht]
	\caption{Summary of 13 Benchmark Functions}
	\label{tableBenchmark}
	\centering
	
	\begin{tabular}{|c|c|c|c|c|c|c|c|}
		\hline
		Fcn & $n$ & Type of $ f $ & $ \rho $ & $LI$ & $NE$ & $NI$ & $ a $ \\
		\hline
		g01 & 13 & quadratic & $ 0.0003\% $ & 9 & 0 & 0 & 6\\
		\hline
		g02 & 20 & nonlinear & $ 99.9965\% $ & 1 & 0 & 1 & 1\\
		\hline
		g03 & 10 & nonlinear & $ 0.0000\% $ & 0 & 1 & 0 & 1\\
		\hline
		g04 & 5 & quadratic & $ 29.9356\% $ & 0 & 0 & 6 & 2\\
		\hline
		g05 & 4 & nonlinear & $ 0.0000\% $ & 2 & 3 & 0 & 3\\
		\hline
		g06 & 2 & nonlinear & $ 0.0064\% $ & 0 & 0 & 2 & 2\\
		\hline
		g07 & 10 & quadratic & $ 0.0003\% $ & 3 & 0 & 5 & 6\\
		\hline
		g08 & 2 & nonlinear & $ 0.8640\% $ & 0 & 0 & 2 & 0\\
		\hline
		g09 & 7 & nonlinear & $ 0.5256\% $ & 0 & 0 & 4 & 2\\
		\hline
		g10 & 8 & linear & $ 0.0005\% $ & 3 & 0 & 3 & 3\\
		\hline
		g11 & 2 & quadratic & $ 0.0000\% $ & 0 & 1 & 0 & 1\\
		\hline
		g12 & 3 & quadratic & $ 0.0197\% $ & 0 & 0 & $ 9^3 $ & 0\\
		\hline
		g13 & 5 & nonlinear & $ 0.0000\% $ & 0 & 3 & 0 & 3\\
		\hline
	\end{tabular}
	
\end{table}%

SMODE contains several parameters which are the population size $ \mu $, the scaling factor $F$ in mutation, the crossover control parameter $Cr$. Usually, $ F $ is set within $ [0,1] $ and mostly from $0.5$ to $0.9$; $ Cr $ is also chosen from $ [0,1] $ and higher values can produce better results in most cases. In our experiments, set $ F $ as 0.6, $ Cr $ as 0.95. The population size $ \mu  =180$.  The tolerance value   $ \delta $ for the equality constraints was set to 0.0001. Set penalty coefficients $c_4=1, c_5=10,c_6=100$. The maximum number of fitness evaluations $FES_{\max}$ is set as $ 5\cdot 10^3 $. 

As suggested in~\cite{liang2006problem}, 25 independent runs are set for each benchmark function.

\subsection{Initial Results of Proposed Algorithm } 

Initial experiments have been completed for $FES_{\max} = 5\cdot 10^3 $ only. 
Table~\ref{table2} shows the result  of function error values achieved  by SMODE with only two helper  functions   $f_1, f_2$  on thirteen benchmark functions.   In the table, NA   means that no feasible solution was found. SMODE may find a feasible solution only on one benchmark function  g06. The result achieved is   worse than that achieved by CMODE  because the infeasible solution replacement mechanism is utilized in SMODE. If this mechanism is added, SMODE is the same as CMODE and their performances could be the same. This is our   ongoing work.

\begin{table*}[ht]
	\caption{Function Error Values Achieved  by SMODE with Only two helper  Functions with FES $=5\cdot 10^3$}
	\label{table2}
	\centering
	
	\begin{tabular}{|c|c|c|c|c|c|c|c|}
		\hline
		Fcn & best & median & worst &　mean distance & Std   \\
		\hline
		g01 & NA &  NA & NA & NA & NA   \\
		\hline
		g02 & NA &  NA & NA & NA & NA   \\
		\hline
		g03 & NA &  NA & NA & NA & NA    \\
		\hline
		g04 & NA &  NA & NA & NA & NA \\
		\hline
		g05 & NA & NA & NA & NA & NA  \\
		\hline
		g06 & 8.9880E+02 & 2.7278E+03 & NA & NA & NA  \\
		\hline
		g07 & NA &  NA & NA & NA & NA  \\
		\hline
		g08 & NA &  NA & NA & NA & NA  \\
		\hline
		g09 & NA &  NA & NA & NA & NA  \\
		\hline
		g10 & NA & NA & NA & NA & NA \\
		\hline
		g11 & NA &  NA & NA & NA & NA  \\
		\hline
		g12  & NA &  NA & NA & NA & NA  \\
		\hline
		g13 & NA & NA & NA & NA & NA  \\
		\hline
	\end{tabular}
	
\end{table*}%

 Table~\ref{table4} gives the result of function error values achieved  by SMODE with four helper functions  $f_1,f_2,f_3, f_4$ on thirteen benchmark functions.    The result achieved by SMODE with four helper functions  is   better than that with only two helper  functions.  SMODE can find feasible solutions on  seven benchmark functions  g2, g4, g6, g8, g9, g11, g12. However, the result achieved is still worse than that achieved by CMODE  because the infeasible solution replacement mechanism is utilized in SMODE. 
 
\begin{table*}[ht]
	\caption{Function Error Values Achieved  by SMODE with 4 help functions   with FES $=5\cdot 10^3$ }
	\label{table4}
	\centering
	
	\begin{tabular}{|c|c|c|c|c|c|c|c|}
		\hline
		Fcn & best & median & worst &　mean distance & Std     \\
		\hline
		g01 & NA & NA & NA & NA & NA  \\
		\hline
		g02 & 5.4500E-01 & 6.1341E-01 & 7.7128E-01 & 8.5337E-01 & 9.4740E-01   \\
		\hline
		g03 & NA & NA & NA & NA & NA    \\
		\hline
		g04 & 3.5549E+02 & 5.9871E+02 & 8.8085E+02 & 1.0152E+02 & 1.3621E+02  \\
		\hline
		g05 & NA & NA & NA & NA & NA  \\
		\hline
		g06 & 3.3688E+02 & 1.3344E+03 & NA & NA & NA  \\
		\hline
		g07 & NA & NA & NA & NA & NA  \\
		\hline
		g08 & 1.1730E-03 & 1.0102E-02 & 5.0376E-02 & 1.1029E-02 & 1.4452E-02  \\
		\hline
		g09 & 8.0239E+01 & 3.1650E+02 & 6.0861E+02 & 1.0339E+02 & 1.2915E+02  \\
		\hline
		g10 & NA & NA & NA & NA & NA   \\
		\hline
		g11 & 1.3360E-03 & 1.2956E-01 & NA & NA & NA   \\
		\hline
		g12 & 6.2614E-05 & 3.9500E-04 & 1.0323E-02 & 2.1810E-03 & 2.8250E-03  \\
		\hline
		g13 & NA & NA & NA & NA & NA  \\
		\hline
	\end{tabular}
	
\end{table*}%

 Table~\ref{table6} is the result of function error values achieved  by SMODE with four helper functions  $f_1,f_2,f_3, f_4$ on thirteen benchmark functions.    The result achieved by SMODE with four helper functions  is similar to that with only six helper functions.  SMODE also can find feasible solutions on  seven benchmark functions  g2, g4, g6, g8, g9, g11, g12. However the difference between four and six helper functions is very small. A possible explanation is that $f_5, f_6$   play a similar rule as $f_4$. All these three functions belong  to the class of penalty functions. Therefore it might be better if helper functions are designed from different backgrounds. 
 
\begin{table*}[htb]
	\caption{Function Error Values Achieved  by SMODE with 6 help functions   with FES $=5\cdot 10^3$  }
	\label{table6}
	\centering
	
	\begin{tabular}{|c|c|c|c|c|c|c|c|}
		\hline
		Fcn & best & median & worst &　mean distance & Std   \\
		\hline
		g01 & NA & NA & NA & NA & NA   \\
		\hline
		g02 & 5.4717E-01 & 5.8388E-01 & 6.2491E-01 & 2.6393E-01 & 3.1753E-01  \\
		\hline
		g03 & NA & NA & NA & NA & NA    \\
		\hline
		g04 & 4.4758E+02 & 6.0336E+02 & 7.7147E+02 & 8.4857E+01 & 9.7692E+01  \\
		\hline
		g05 & NA & NA & NA & NA & NA  \\
		\hline
		g06 & 4.1811E+02 & 2.8303E+03 & 5.1241E+02 & 1.1938E+03 & 1.4209E+03 \\
		\hline
		g07 & NA & NA & NA & NA & NA  \\
		\hline
		g08 & 1.4000E-05  & 2.6850E-03 & 2.0954E-02 & 4.4980E-03 & 5.9500E-03  \\
		\hline
		g09 & 1.0152E+02 & 3.4975E+02 & 9.3096E+02 & 9.2581E+01 & 1.5020E+02  \\
		\hline
		g10 & NA & NA & NA & NA & NA  \\
		\hline
		g11 & 1.2160E-03 & 1.3616E-01 & NA & NA & NA  \\
		\hline
		g12 & 5.4000E-05 & 2.6000E-04 & 1.0015E-02 & 2.7600E-03 & 3.2760E-03  \\
		\hline
		g13  & NA & NA & NA & NA & NA  \\
		\hline
	\end{tabular}
	
\end{table*}%

In summary, our initial experimental results confirm that the performance of SMODE with more helper functions (four or six) is better than that of  that with only two helper  functions. Currently SMODE performs worse than CMODE. But if the infeasible solution replacement mechanism is added to SMODE, SMODE is the same as CMODE  and it is expected that their performances could be the same.

\section{Conclusion and Future Work}
\label{secConclusion}
This paper proposes  a new multi-objective method for solving constrained optimization problems. 
The new method  keeps two standard    objectives:  the objective function and also the sum of degrees of constraint violation. But besides them, four more objectives are added. One is based on the feasible rule. The   other three come from the penalty functions.   

This paper conducts an initial experimental study on thirteen benchmark functions. A simplified version of CMODE~\cite{wang2012combining} is applied to solving    multi-objective optimization problems. Our initial experimental results are positive. They confirm our expectation that adding helper functions could be useful. The performance of SMODE with more helper functions (four or six) is better than that with only two helper  functions.

Due to time limitation, a key part in CMODE, the infeasible solution replacement mechanism, is  not implemented in SMODE. Thus  the result achieved by SMODE is worse than that achieved by CMODE. But if this mechanism is added to SMODE, SMODE is the same as CMODE and their performances could be the same. A study on the original CMODE with different numbers of helper functions is our   ongoing work.




%\bibliographystyle{IEEEtran}

\end{document}